\documentclass[preprint,12pt]{elsarticle}

\newif\ifblindreview
\blindreviewfalse

\usepackage[hidelinks]{hyperref}
\hypersetup{
  pdfauthor={\ifblindreview Anonymous \else Ashish Jha; Valentin Leplat; Anh Huy Phan \fi},
  pdftitle={Market-Driven Subset Selection for Budgeted Training}
}

\usepackage{amssymb}
\usepackage{amsmath}
\usepackage{amsthm}
\usepackage{graphicx}
\usepackage{booktabs}
\usepackage{multirow}
\usepackage{algorithm}
\usepackage{algpseudocode}
\usepackage{tabularx}
\usepackage{array}
\usepackage{xcolor}
\usepackage{hyperref}
\usepackage{booktabs}
\usepackage{siunitx}
\usepackage[table]{xcolor}
\usepackage[normalem]{ulem} % for \uline (underline)
\newcommand{\best}[1]{\textbf{#1}}
\newcommand{\second}[1]{\uline{#1}}
\newcommand{\pos}[1]{\cellcolor{green!10}{#1}}
\renewcommand{\neg}[1]{\cellcolor{red!8}{#1}}
\newcommand{\neu}[1]{#1}

\theoremstyle{plain}
\newtheorem{theorem}{Theorem}[section]
\newtheorem{proposition}[theorem]{Proposition}

\newtheorem{lemma}[theorem]{Lemma}

\theoremstyle{definition}

\theoremstyle{remark}

\journal{Information Processing \& Management}

\begin{document}

\begin{frontmatter}

\title{Market-Driven Subset Selection for Budgeted Training}

\ifblindreview
  % ---- Anonymous version ----
  \author{Anonymous}
  \address{Affiliations withheld for double-blind review}
  % Do NOT include \ead, \cortext, ORCID, grants, acknowledgments, or data/code links that reveal identity.
\else
  % ---- Camera-ready version ----
  \author[skoltech]{Ashish Jha}
  \ead{ashish.jha@skoltech.ru}

  \author[innopolis]{Valentin Leplat}
  \ead{v.leplat@innopolis.ru}

  \author[skoltech]{Anh Huy Phan\corref{cor1}}
  \ead{a.phan@skoltech.ru}
  \cortext[cor1]{Corresponding author}

  \affiliation[skoltech]{organization={Skolkovo Institute of Science and Technology},
              addressline={Bolshoy Boulevard 30, bld. 1},
              city={Moscow},
              postcode={121205},
              country={Russian Federation}}

  \affiliation[innopolis]{organization={Innopolis University},
              addressline={Universitetskaya St., 1},
              city={Innopolis},
              postcode={420500},
              state={Republic of Tatarstan},
              country={Russian Federation}}
\fi
\begin{abstract}
Training large language models on massive datasets is computationally expensive, yet empirical evidence suggests that substantial portions of training examples contribute minimally to final performance. Data subset selection addresses this inefficiency by identifying small, high-utility subsets under resource constraints. However, example utility is inherently multi-faceted, encompassing uncertainty, distributional rarity, and diversity signals that are heterogeneous and typically combined through ad hoc weighted sums lacking theoretical grounding. We propose a market-based framework that treats each training example as a tradeable contract and employs the Logarithmic Market Scoring Rule to aggregate multiple utility signals into coherent prices. Heterogeneous signals act as traders, a single liquidity parameter controls concentration versus smoothing, and topic-wise normalization ensures calibrated aggregation. Token budgets are handled explicitly through a price-per-token decision rule with an interpretable length-bias parameter. We establish theoretical connections to maximum-entropy aggregation and provide utility recovery guarantees under noisy but monotone signals. On GSM8K mathematical reasoning under strict 60k-token budgets, our selector achieves parity with strong single-signal baselines while exhibiting lower variance and incurring less than 0.1 GPU-hour overhead. On AGNews classification at 5--25\% retention rates, the market formulation delivers competitive accuracy with improved stability. Our framework unifies multi-signal data curation under fixed computational budgets for prompt-level reasoning and classification tasks.
\end{abstract}

% \begin{highlights}
% \item Novel market-based framework treating data selection as prediction market pricing via LMSR
% \item Principled aggregation of heterogeneous utility signals with maximum-entropy guarantees
% \item Token-aware decision rules with explicit length-bias for variable-cost data items
% \item Competitive performance with improved stability on reasoning and classification benchmarks
% \item Computational efficiency with sub-0.1 GPU-hour selection overhead
% \end{highlights}


\end{frontmatter}

\section{Introduction}
\label{sec:introduction}

The computational demands of training large language models on ever-expanding datasets have made data efficiency a central concern in modern machine learning. While larger datasets generally improve model capabilities, substantial evidence indicates that many training examples contribute little to final performance \cite{toneva2019forgetting,paul2021deep}. This observation motivates data subset selection: identifying a small, high-utility subset that preserves performance while operating under fixed computational budgets.

The fundamental challenge lies in the heterogeneous nature of example utility. Model uncertainty on an example, its distributional rarity, diversity relative to already-selected examples, and task-specific difficulty signals often provide conflicting recommendations. Existing approaches address this heterogeneity through either single-signal optimization, such as pure uncertainty sampling \cite{settles2009active}, or ad hoc weighted combinations that lack theoretical justification and exhibit brittleness across tasks and domains. This challenge intensifies in instruction tuning and prompt-level reasoning for large language models, where individual examples have highly variable lengths and computational costs. A 10-token arithmetic question and a 500-token reasoning problem both constitute single training examples yet represent vastly different computational investments. Traditional cardinality-based selection methods fail to account for this cost heterogeneity.

We propose casting data subset selection as a prediction market where each training example is a contract whose price reflects its aggregated utility. We employ the Logarithmic Market Scoring Rule, a well-studied market-making mechanism from economics and information aggregation \cite{hanson2007lmsr}, to convert heterogeneous utility signals into coherent probability distributions over examples. Our framework provides principled aggregation with a single liquidity parameter controlling concentration versus smoothing, explicit handling of variable token costs through an interpretable length-bias parameter, and computational efficiency suitable for large-scale applications.
% After the last paragraph of the Introduction
\begin{figure*}[t]
  \centering
  \includegraphics[width=\textwidth]{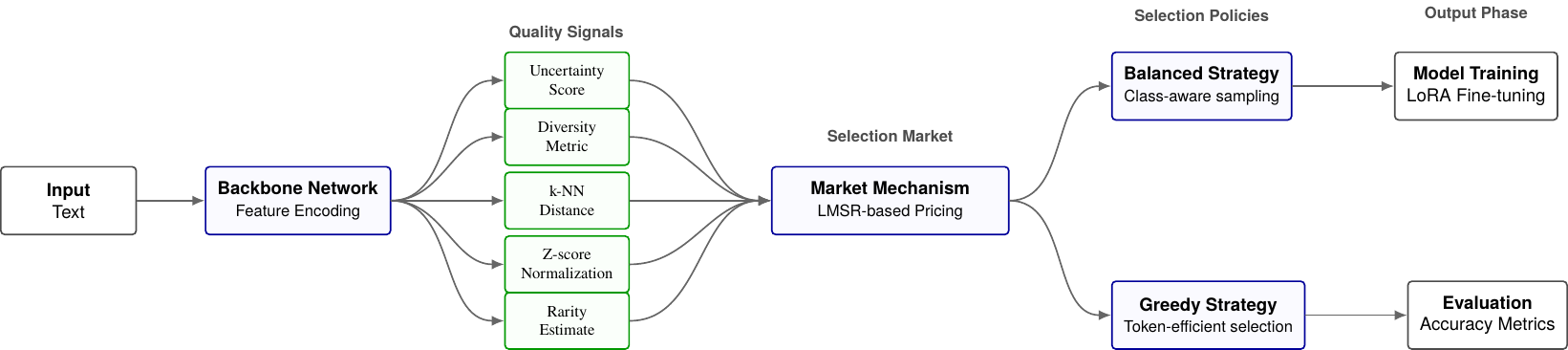}
  \caption{End-to-end market-based data selection pipeline:
  heterogeneous signals $\to$ LMSR pricing $\to$ token-aware selection $\to$
  training and evaluation.}
  \label{fig:pipeline}
\end{figure*}

\subsection{Contributions}

This work introduces a market-based aggregation framework for data selection. We formulate selection as pricing in a cost-function prediction market where utility signals act as traders and LMSR provides principled aggregation. The framework handles variable example costs through a token-aware decision rule that ranks examples by price-per-token adjusted by an interpretable length-bias exponent. We establish theoretical connections showing that LMSR implements maximum-entropy aggregation with exponential weighting, yielding transparent control over aggregation strength. Under assumptions of weakly informative but monotone signals, we provide utility recovery guarantees showing that exponential weighting amplifies aligned signals while averaging independent noise.

Empirically, we evaluate on mathematical reasoning tasks under strict token budgets and text classification at multiple retention rates. On GSM8K with 60k-token budgets, the market selector achieves robust parity with strong single-signal baselines while reducing variance across random seeds. On AGNews classification, market-based selection with light balancing delivers competitive accuracy with improved stability. The selection overhead remains below 0.1 GPU-hour, making the approach practical for large-scale data curation. Our framework unifies multi-signal aggregation for both prompt-level reasoning and classification under fixed computational constraints.

\subsection{Paper Organization}

Section \ref{sec:background} reviews related work on data subset selection, active learning, and information aggregation through prediction markets. Section \ref{sec:method} presents the market-based framework, including LMSR pricing, signal aggregation, and token-aware selection rules. Section \ref{sec:theory} provides theoretical analysis connecting LMSR to maximum-entropy aggregation and establishing utility recovery guarantees. Section \ref{sec:signals} details the construction and standardization of utility signals for language model training. Section \ref{sec:experiments} presents experimental validation on reasoning and classification benchmarks. Section \ref{sec:discussion} discusses limitations and practical considerations. Section \ref{sec:conclusion} concludes with future directions.

\section{Background and Related Work}
\label{sec:background}

Data subset selection addresses the question of identifying informative examples from large pools while respecting computational constraints. We review relevant work across geometric coverage methods, gradient-based selection, uncertainty and diversity criteria from active learning, and the use of prediction markets for information aggregation.

\subsection{Geometric Coverage and Coresets}

Geometric approaches to data selection frame the problem as finding representative subsets that preserve structural properties of the full dataset. Coreset construction methods aim to find small weighted subsets such that model parameters trained on the coreset approximate those trained on the full data \cite{mirzasoleiman2020coresets}. Facility location and $k$-center objectives provide coverage guarantees by ensuring that every point in the original set lies close to some selected point \cite{sener2018active}. These methods typically optimize submodular objectives that exhibit diminishing returns, enabling approximation guarantees through greedy algorithms \cite{nemhauser1978submodular}. However, most coreset approaches assume uniform example costs and do not naturally extend to settings where examples have variable computational requirements.

\subsection{Gradient-Based and Bilevel Selection}
Gradient-based methods select examples that align with or accelerate optimization objectives.
GLISTER~\cite{killamsetty2021glister} formulates selection as a bilevel optimization problem where the inner loop trains on a candidate subset and the outer loop optimizes subset composition to minimize validation loss.
Gradient matching approaches select examples whose gradients align with those computed on validation data.
While these methods can integrate multiple implicit criteria through the training objective, they require expensive validation-set evaluations at each selection iteration and lack interpretable control parameters for practitioners to adjust selection behavior. More recently, GRAFT~\cite{jha2025graft} proposes a gradient-aware, fast MaxVol-style selector that prioritizes samples with high agreement and coverage in gradient space, substantially reducing selection overhead compared to full gradient-matching.
In contrast, our market-based approach aggregates heterogeneous non-gradient signals (loss/uncertainty, rarity, length) via an LMSR scoring rule, yielding interpretable knobs ($\beta,\gamma$) and plug-in simplicity without gradient computations or validation passes.

\subsection{Active Learning and Uncertainty Sampling}

Active learning provides a rich framework for iterative data selection based on model uncertainty. Uncertainty sampling selects examples where the model is least confident \cite{settles2009active,lewis1994uncertainty}. Bayesian approaches such as BALD \cite{houlsby2011bald} and BatchBALD \cite{kirsch2019batchbald} measure information gain about model parameters. Entropy-based methods select examples with high predictive entropy. Diversity-aware active learning combines uncertainty with geometric diversity \cite{ash2020badge}, often through clustering or determinantal point processes. However, most active learning methods optimize a single primary criterion, such as uncertainty, and add diversity as a secondary constraint rather than providing principled multi-criteria aggregation.

\subsection{Training Dynamics and Forgetting}

Recent work exploits training dynamics to identify influential or difficult examples. Forgetting events track how often examples transition from correctly classified to misclassified during training \cite{toneva2019forgetting}, with frequently forgotten examples indicating challenging learning signals. The Error L2-Norm (EL2N) metric measures the L2 norm of prediction errors early in training \cite{paul2021deep}, serving as a proxy for example difficulty. Gradient norms (GraNd) identify examples with large gradient magnitudes. These dynamics-based scores provide complementary information to static uncertainty or geometric diversity, yet they are typically used in isolation or combined through ad hoc weighting schemes.

\subsection{Prediction Markets for Information Aggregation}

Prediction markets aggregate information from multiple sources by allowing participants to trade contracts whose payoffs depend on future outcomes. The Logarithmic Market Scoring Rule provides a market-making mechanism with desirable properties including proper scoring, convexity, and bounded loss \cite{hanson2007lmsr}. LMSR has been studied extensively in mechanism design and online learning \cite{abernethy2013efficient,chen2010noregret}. Recent work applies market-based aggregation to ensemble predictions and crowdsourcing \cite{lambert2008scoring}. However, to our knowledge, using LMSR to price data examples for subset selection represents a novel application of prediction market theory to training data curation.

\subsection{Positioning of Our Work}

Our framework differs from prior work in several key aspects. Unlike geometric coverage methods, we provide principled multi-signal aggregation rather than optimizing a single submodular objective. Unlike bilevel optimization approaches, we offer interpretable control parameters and avoid expensive nested optimization loops. Unlike active learning methods that prioritize a single criterion with diversity as a constraint, we treat all signals symmetrically through the market mechanism. Unlike dynamics-based scores used in isolation, we provide a theoretically grounded aggregation method. Our use of LMSR connects data selection to the rich theory of prediction markets, offering maximum-entropy characterization and convexity guarantees absent from ad hoc weighted combinations.

\section{Market-Based Data Selection Framework}
\label{sec:method}

We present the market-based data selection framework, beginning with the formulation of selection as market pricing, followed by the aggregation of heterogeneous signals, token-aware decision rules, and topic-separable markets for handling heterogeneous data groups.

\subsection{Subset Selection as Market Pricing}

Let $\mathcal{D}=\{(x_i,y_i)\}_{i=1}^N$ denote a pool of training examples, and let $\mathcal{S}\subseteq\mathcal{D}$ denote a selected subset. We cast selection as pricing in a cost-function prediction market where each example $i$ is a contract with price $p_i$ reflecting its utility for training. The Logarithmic Market Scoring Rule defines a convex cost function over outstanding shares $q\in\mathbb{R}^N$ with liquidity parameter $\beta>0$:
\begin{equation}
\label{eq:lmsr-cost}
C(q) = \beta \log \!\Big(\sum_{j=1}^N e^{\,q_j/\beta}\Big).
\end{equation}
Prices are defined as the gradient of the cost function:
\begin{equation}
\label{eq:lmsr-price}
p_i(q) = \frac{\partial C}{\partial q_i}
= \frac{e^{\,q_i/\beta}}{\sum_j e^{\,q_j/\beta}}
= \big[\mathrm{softmax}(q/\beta)\big]_i.
\end{equation}

\begin{figure}[t]
  \centering
  \includegraphics[width=\linewidth]{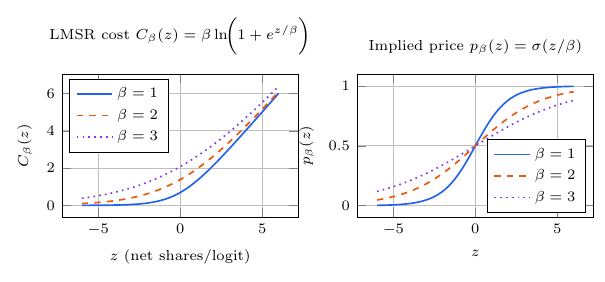}
  \caption{LMSR mechanics: (left) cost $C_\beta(q)$; (right) implied prices
  $p(q)=\mathrm{softmax}(q/\beta)$. Larger $\beta$ flattens curvature and smooths prices.}
  \label{fig:lmsr-mechanics}
\end{figure}

Consequently, $p\in\Delta^{N-1}$ forms a probability distribution over examples. The liquidity parameter $\beta$ controls the concentration of prices: smaller $\beta$ yields sharper concentration on high-share examples, while larger $\beta$ produces smoother distributions.

% Immediately after the paragraph that introduces LMSR + beta
\begin{figure}[t]
  \centering
  \includegraphics[width=\linewidth]{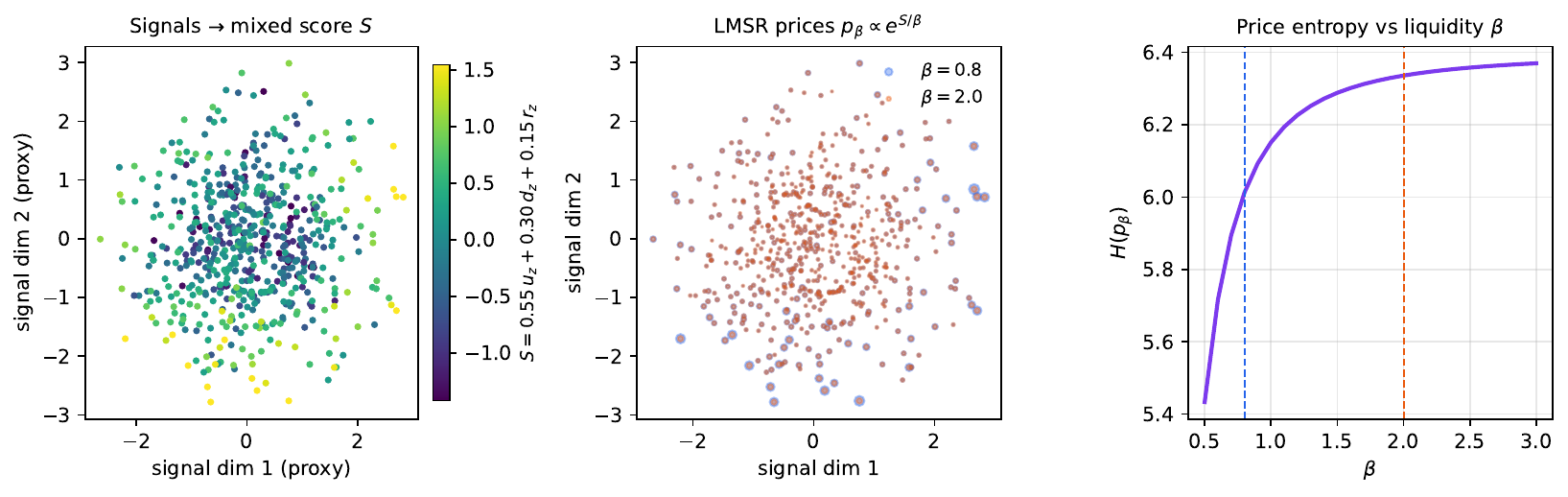}
  \caption{LMSR-based aggregation of heterogeneous signals. Liquidity $\beta$
  controls smoothing of prices, interpolating between concentrated and diffuse mixes.}
  \label{fig:lmsr-aggregation}
\end{figure}

Figure \ref{fig:lmsr-aggregation} illustrates this behavior. As $\beta$ decreases, prices concentrate sharply on examples with the highest shares, implementing a winner-take-all dynamic. As $\beta$ increases, prices smooth toward uniform distribution, giving more weight to lower-ranked examples. This temperature-like behavior provides interpretable control over aggregation strength.

The LMSR framework inherits desirable properties from its foundations in proper scoring rules and convex analysis. The cost function is strictly convex, ensuring unique equilibrium prices for any share vector. The pricing rule satisfies the proper scoring property, meaning that truthful reporting of beliefs maximizes expected utility for rational traders. These properties have been established in the context of prediction markets and online learning \cite{hanson2007lmsr,abernethy2013efficient,chen2010noregret}.

\subsection{Aggregating Heterogeneous Signals}

Example utility for machine learning is inherently multi-faceted. A mathematically challenging problem may be distributionally common, while a rare example may be easy for the current model. Diversity relative to already-selected examples provides yet another dimension orthogonal to both difficulty and rarity. Our framework aggregates these heterogeneous signals by treating each signal as a trader contributing to the share vector.

Let $s^{(m)}_i$ denote the raw value of signal $m$ for example $i$, where $m$ ranges over available signals such as uncertainty, rarity, and diversity. We standardize each signal within topic groups (discussed in Section \ref{sec:topic-markets}) to obtain normalized scores $\tilde{s}^{(m)}_i$. The aggregate share for example $i$ is computed as a weighted linear combination:
\begin{equation}
\label{eq:shares}
q_i = \sum_{m=1}^M w_m \tilde{s}^{(m)}_i,
\end{equation}
where $w_m \geq 0$ are signal weights. In the simplest case, we use equal weights $w_m = 1/M$. When a small development set is available, we can tune weights via online learning methods such as multiplicative weights or exponentiated gradient descent \cite{arora2012mwu}.

After computing shares, prices follow from Equation \ref{eq:lmsr-price}. This two-stage process separates the task of measuring diverse aspects of utility (signal computation and standardization) from the task of aggregating them into a coherent ranking (LMSR pricing). The separation provides modularity: new signals can be added by extending the sum in Equation \ref{eq:shares} without changing the pricing mechanism.

\subsection{Token-Aware Selection Under Budget Constraints}

Modern language model training operates under token-level budgets rather than example-count budgets. A dataset of instruction-following examples may contain short single-turn dialogues of 50 tokens alongside multi-turn conversations exceeding 1000 tokens. Selecting a fixed number of examples without considering length can lead to inefficient budget allocation.

We address this by ranking examples using a price-per-token score adjusted for length bias. For example $i$ with token length $\ell_i$ and price $p_i$, define the selection score:
\begin{equation}
\label{eq:price-per-token}
\rho_i = \frac{p_i}{\ell_i^\gamma},
\end{equation}
where $\gamma \geq 0$ is an interpretable length-bias exponent. When $\gamma=0$, selection reduces to ranking by raw price, favoring high-utility examples regardless of length. When $\gamma=1$, selection ranks by price-per-token, treating tokens as the fundamental unit. Intermediate values interpolate between these extremes.

The length-bias parameter provides practitioners with explicit control over the trade-off between selecting many short examples versus fewer long examples. In practice, we find that $\gamma \in [1.4, 1.8]$ works well across diverse tasks, with $\gamma=1.6$ serving as a robust default. After computing $\rho_i$ for all examples, we perform greedy selection: sort examples in descending order of $\rho_i$ and add examples to the selected set until the cumulative token count reaches the budget $B$.

\subsection{Topic-Separable Markets}
\label{sec:topic-markets}

Real datasets often contain heterogeneous subgroups or topics. In mathematical reasoning datasets, problems may span arithmetic, algebra, geometry, and combinatorics. In instruction-following datasets, examples may cover different task types such as summarization, question answering, and creative writing. Without accounting for this heterogeneity, signals from high-volume topics can dominate the aggregated shares, leading to poor coverage of minority topics.

We address this through topic-separable LMSR markets. Partition the examples into topic groups $\mathcal{I}_t$ for $t \in \mathcal{T}$, where $t(i)$ denotes the topic of example $i$. Introduce topic budgets $\alpha_t \geq 0$ satisfying $\sum_{t \in \mathcal{T}} \alpha_t = 1$, representing the desired fraction of total probability mass allocated to topic $t$. The topic-separable cost function is:
\begin{equation}
\label{eq:topic-cost}
C(q) = \sum_{t\in\mathcal{T}} \alpha_t\,\beta_t \log\!\Big(\sum_{j\in \mathcal{I}_t} e^{\,q_j/\beta_t}\Big).
\end{equation}
Prices are obtained by taking gradients:
\begin{equation}
\label{eq:topic-price}
p_i = \frac{\partial C}{\partial q_i} = \alpha_{t(i)}\Big[\mathrm{softmax}_{\mathcal{I}_{t(i)}}\!\big(q/\beta_{t(i)}\big)\Big]_i.
\end{equation}
This formulation ensures that example $i$ in topic $t$ receives price proportional to $\alpha_t$ and competes only against other examples within the same topic via the topic-local softmax. The global price vector $p$ remains a probability distribution with $\sum_i p_i = 1$.

Signal standardization is performed within topics. For signal $m$ and topic $t$, compute the topic-conditional mean $\mu_t^{(m)}$ and standard deviation $\sigma_t^{(m)}$ over examples in $\mathcal{I}_t$. The standardized signal for example $i \in \mathcal{I}_t$ is:
\begin{equation}
\tilde{s}^{(m)}_i = \frac{s^{(m)}_i - \mu_t^{(m)}}{\sigma_t^{(m)}}.
\end{equation}
For heavy-tailed distributions, we apply rank-based normalization or robust scaling using median and interquartile range before computing shares.

The topic-separable formulation provides fairness guarantees across heterogeneous subgroups while preserving within-group competition. If topic budgets are set proportional to topic size, $\alpha_t \propto |\mathcal{I}_t|$, the market preserves the original topic distribution. If budgets are tuned on a development set, the market can shift capacity toward higher-yield topics while maintaining interpretable control through the $\alpha_t$ parameters.

\subsection{Algorithm and Computational Complexity}

Algorithm \ref{alg:lmsr_market} summarizes the market-based selection procedure. The algorithm first computes utility signals for all examples and standardizes them within topics. Shares are computed as weighted combinations of standardized signals. Prices are obtained via topic-separable LMSR. Finally, examples are ranked by price-per-token score and greedily selected until the token budget is exhausted.

\begin{algorithm}[t]
\caption{Market-Based Data Selection}
\label{alg:lmsr_market}
\small
\begin{algorithmic}[1]
\Require Dataset $\mathcal{D}$; topics $t(i)\!\in\!\mathcal{T}$; signals $\{\phi^{(k)}\}_{k=1}^K$; token budget $B$; topic budgets $\{\alpha_t\}$; liquidities $\{\beta_t\}$; length-bias $\gamma$; signal weights $\{w_k\}$
\Ensure Selected subset $\mathcal{S}\subseteq\mathcal{D}$ with token cost $\leq B$
\State Compute raw signals: $s^{(k)}_i \gets \phi^{(k)}(x_i,y_i)$ for all $i,k$
\State Standardize within topics: $\tilde{s}^{(k)}_i \gets (s^{(k)}_i - \mu_{t(i)}^{(k)})/\sigma_{t(i)}^{(k)}$
\State Compute shares: $q_i \gets \sum_{k=1}^K w_k\, \tilde{s}^{(k)}_i$ for all $i$
\State Compute prices: $p_i \gets \alpha_{t(i)} \cdot [\mathrm{softmax}_{\mathcal{I}_{t(i)}}(q/\beta_{t(i)})]_i$ for all $i$
\State Compute selection scores: $\rho_i \gets p_i / \ell_i^{\gamma}$ for all $i$
\State Sort examples by $\rho_i$ in descending order
\State $\mathcal{S} \gets \emptyset$; $\text{tokens}_\text{used} \gets 0$
\For{$i$ in sorted order}
    \If{$\text{tokens}_\text{used} + \ell_i \leq B$}
        \State $\mathcal{S} \gets \mathcal{S} \cup \{i\}$
        \State $\text{tokens}_\text{used} \gets \text{tokens}_\text{used} + \ell_i$
    \EndIf
\EndFor
\State \Return $\mathcal{S}$
\end{algorithmic}
\end{algorithm}

The computational complexity of the algorithm is modest. Signal computation depends on the specific signals used; uncertainty-based signals require a forward pass through the model, costing $O(NL)$ where $L$ is average sequence length. Diversity and rarity signals based on approximate nearest neighbors incur $O(N\log N)$ cost for index construction and $O(N\log N)$ for queries using algorithms such as HNSW \cite{malkov2018hnsw}. Standardization and share computation are $O(NK)$ where $K$ is the number of signals. LMSR pricing is $O(N)$ for computing the softmax. Sorting for greedy selection is $O(N\log N)$. The dominant cost is typically the forward pass for uncertainty signals or the ANN index for diversity signals, both of which are standard operations in data selection pipelines. In our experiments, the end-to-end selection overhead remains below 0.1 GPU-hour even for datasets with tens of thousands of examples.

\section{Theoretical Analysis}
\label{sec:theory}

We establish theoretical properties of the market-based framework, showing that LMSR implements maximum-entropy aggregation and providing utility recovery guarantees under noisy but informative signals.

\subsection{Maximum-Entropy Characterization}

The LMSR pricing mechanism admits an elegant information-theoretic interpretation as maximum-entropy aggregation. This connection clarifies why the market formulation provides principled multi-signal integration.

\begin{proposition}[Maximum-Entropy Aggregation]
\label{prop:maxent}
Let $s^{(m)}_i$ denote signal $m$ evaluated on example $i$. Consider the optimization problem:
\begin{equation}
\label{eq:maxent-prob}
\max_{p\in\Delta^{N-1}} H(p) \quad \text{subject to} \quad \mathbb{E}_{p}[s^{(m)}] = \bar{s}^{(m)} \quad \forall m,
\end{equation}
where $H(p) = -\sum_i p_i \log p_i$ is the Shannon entropy and $\bar{s}^{(m)}$ are target moments. The unique solution has the exponential form:
\begin{equation}
p_i^\star \propto \exp\!\Big(\sum_{m=1}^M \lambda_m s^{(m)}_i\Big),
\end{equation}
where $\lambda_m$ are Lagrange multipliers. Setting $\lambda_m = w_m/\beta$ recovers LMSR prices from Equation \ref{eq:lmsr-price}.
\end{proposition}

The proof follows from standard convex duality for entropy maximization with linear constraints \cite{jaynes1957information}. The entropy functional is strictly concave, and the feasible set defined by the moment constraints is convex, ensuring uniqueness of the solution. The Lagrange multipliers arise as dual variables, yielding the exponential family form. The LMSR cost function in Equation \ref{eq:lmsr-cost} serves as the log-partition function of this exponential family, with prices given by its gradient.

This characterization provides several insights. First, the market mechanism chooses the least committal probability distribution consistent with the observed signals, embodying the principle of maximum entropy. Second, the liquidity parameter $\beta$ controls the inverse temperature of the exponential family, interpolating smoothly between concentrated and diffuse distributions. Third, the aggregation respects the convex geometry of probability distributions, unlike ad hoc weighted sums that may produce scores outside meaningful ranges.

\subsection{Utility Recovery Under Noisy Signals}

In practice, utility signals are imperfect proxies for true example utility. Uncertainty estimates may be miscalibrated, diversity scores may be affected by embedding quality, and rarity measures depend on density estimation in high dimensions. We analyze the robustness of LMSR aggregation under signal noise.

Assume that true example utilities $U_i \in [0,1]$ exist but are unobserved. Each signal $s^{(m)}_i$ provides a noisy estimate of utility, satisfying:
\begin{equation}
s^{(m)}_i = f_m(U_i) + \epsilon^{(m)}_i,
\end{equation}
where $f_m$ are monotone functions and $\epsilon^{(m)}_i$ are zero-mean sub-Gaussian noise terms with variance $\sigma^2_m$. After standardization and aggregation, the combined score is:
\begin{equation}
q_i = \sum_{m=1}^M w_m \tilde{s}^{(m)}_i.
\end{equation}

\begin{proposition}[Utility Recovery with Weak Signals]
\label{prop:utility-recovery}
Assume that each signal function $f_m$ is monotone increasing in true utility $U$ and that noise terms $\epsilon^{(m)}_i$ are independent sub-Gaussian with parameter $\sigma$. Let $\Gamma$ denote the minimum signal-to-noise ratio across signals. Select the top-$K$ examples by LMSR prices $p_\beta$. Then with probability at least $1-\delta$, the selected set achieves total utility at least $(1-\varepsilon)\,\mathrm{OPT}_K$, where:
\begin{equation}
\varepsilon = O\!\Big(\sigma \sqrt{\frac{\log(N/\delta)}{K\,\Gamma^2}}\Big).
\end{equation}
\end{proposition}

The proof follows from concentration inequalities for sums of sub-Gaussian random variables. The exponential weighting in LMSR amplifies the aligned signal component while independent noise terms average toward zero. The bound shows that the approximation error decreases with the square root of the budget size $K$ and the squared signal-to-noise ratio $\Gamma$. Stronger signals or larger budgets improve recovery guarantees.

This result clarifies the role of exponential weighting in robust aggregation. Unlike arithmetic means that give equal weight to all signals, exponential weighting naturally upweights signals with stronger alignment to true utility. This aligns with intuitions from boosting and multiplicative weights algorithms in online learning \cite{arora2012mwu}, where exponential reweighting amplifies informative features.

\subsection{Coverage and Diversity Guarantees}

Beyond selecting high-utility examples, practical data selection must ensure coverage of diverse example types. We analyze coverage properties when diversity signals receive positive weight in the market.

\begin{proposition}[Coverage with Diversity Signal]
\label{prop:coverage}
Suppose a diversity signal based on centroid distance or $k$-nearest-neighbor anti-density receives weight $w_\text{div} > 0$ in the share computation. Then the selected subset $\mathcal{S}$ with $|\mathcal{S}|=K$ satisfies:
\begin{equation}
\text{Var}_\mathcal{S}[\phi] \geq c \cdot \text{Var}_\mathcal{D}[\phi] \quad \text{and} \quad r_\text{cover}(\mathcal{S}) = O\!\Big(\sqrt{\frac{d\log(N/\delta)}{K}}\Big),
\end{equation}
where $\phi$ denotes the embedding function, $c>0$ is a constant depending on $w_\text{div}$, and $r_\text{cover}$ is the covering radius with probability $1-\delta$.

\end{proposition}

This proposition connects our market-based approach to classical covering and facility-location objectives. When diversity receives non-trivial weight, the selected subset preserves a constant fraction of the variance in the embedding space and achieves covering radius comparable to $k$-center algorithms \cite{gonzalez1985clustering}. The bound on covering radius matches the minimax rate for metric covering problems, confirming that adding a diversity signal to the market does not incur substantial coverage penalties relative to pure geometric methods.

\subsection{Robustness to Signal Corruption}

Real-world signals may be corrupted by calibration errors, adversarial manipulation, or distribution shift. We analyze the sensitivity of LMSR prices to corruption in a single signal.

\begin{lemma}[Bounded Influence Under Corruption]
\label{lem:bounded-influence}
Suppose all signals are clipped to $[-\tau, \tau]$ after standardization. Consider corruption of signal $m^\star$ by replacing $z_{i,m^\star}$ with $(1-\varepsilon)z_{i,m^\star} + \varepsilon\eta_i$ where $|\eta_i|\leq\tau$ and $\varepsilon\in[0,1]$. Let $p$ and $p'$ denote prices before and after corruption. Then:
\begin{equation}
\|p'-p\|_1 \leq 2\,\beta\,\tau\,\varepsilon\,w_{m^\star}.
\end{equation}
\end{lemma}

The proof exploits the Lipschitz property of the softmax function. The perturbation in shares is bounded by $2\tau\varepsilon w_{m^\star}$ due to clipping, and the gradient of the softmax has operator norm at most $\beta$. This result shows that the impact of a corrupted signal scales linearly with the corruption level $\varepsilon$, the clipping radius $\tau$, the signal weight $w_{m^\star}$, and the liquidity $\beta$. Practitioners can control sensitivity by choosing modest clipping thresholds and appropriate liquidity values.

\section{Utility Signals for Language Model Training}
\label{sec:signals}

We detail the construction of utility signals for instruction tuning and prompt-level reasoning tasks. Each signal captures a complementary aspect of example utility, and all signals are standardized within topics before aggregation.

\subsection{Uncertainty-Based Signals}

Uncertainty signals measure how confident the model is on each example. For instruction-following tasks, we compute the mean token-level negative log-likelihood of the target response given the instruction under a base or teacher language model. For example $(x_i, y_i)$ where $x_i$ is the instruction and $y_i$ is the target response of length $\ell_i$ tokens, the uncertainty score is:
\begin{equation}
s_i^\text{unc} = -\frac{1}{\ell_i}\sum_{t=1}^{\ell_i} \log p_\theta(y_{i,t} \mid x_i, y_{i,<t}),
\end{equation}
where $\theta$ denotes the base model parameters. Length normalization ensures that long responses do not dominate purely due to their length. Higher uncertainty indicates examples where the model is less confident, suggesting potential learning value.

For chain-of-thought reasoning tasks, we augment token-level uncertainty with self-consistency dispersion. We sample multiple reasoning chains from the model and measure the variance in final answers or intermediate reasoning steps. Examples where the model produces inconsistent reasoning receive higher uncertainty scores, indicating ambiguity that training may resolve.

For code generation and mathematical problem solving, we incorporate correctness proxies when ground-truth evaluations are available. We score the next-token distribution restricted to answer formats, such as boxed numerals for math problems or unit-tested code stubs. When execution environments exist, we augment uncertainty with binary success indicators from running generated code against test cases.

\subsection{Rarity and Distributional Coverage}

Rarity signals identify examples that are unusual or under-represented in the training pool. We compute rarity in embedding space using $k$-nearest-neighbor anti-density. For each example $i$, extract a representation $\phi(x_i)$ using a pretrained sentence encoder. Compute the average distance to the $k$ nearest neighbors:
\begin{equation}
s_i^\text{rare} = \frac{1}{k}\sum_{j\in\text{kNN}(i)} \|\phi(x_i) - \phi(x_j)\|_2.
\end{equation}
Examples with large $k$-NN distances lie in low-density regions of the embedding space, suggesting distributional rarity. We use approximate nearest neighbor indices such as FAISS or HNSW for efficient computation at scale \cite{johnson2019faiss,malkov2018hnsw}.

Rarity is computed within topics to avoid confounding topic identity with distributional coverage. An example that is common within its topic but belongs to a rare topic should not receive inflated rarity scores. Topic-conditional rarity ensures that we identify genuinely unusual examples within each semantic cluster.

\subsection{Diversity Signals}

Diversity signals discourage redundancy in the selected subset. We construct diversity scores by combining centroid distance with $k$-NN anti-density. After clustering examples by topic or semantic similarity, compute the centroid of each cluster. For example $i$ in cluster $c$, the centroid distance is:
\begin{equation}
s_i^\text{div-cent} = \|\phi(x_i) - \mu_c\|_2,
\end{equation}
where $\mu_c$ is the cluster centroid. Examples far from their cluster centroid contribute more to geometric coverage.

We augment centroid distance with local anti-density to handle isolated examples. The combined diversity score is:
\begin{equation}
s_i^\text{div} = \alpha_\text{cent} \cdot s_i^\text{div-cent} + \alpha_\text{knn} \cdot s_i^\text{rare},
\end{equation}
where $\alpha_\text{cent}$ and $\alpha_\text{knn}$ are fixed combination weights. This formulation rewards both distance from cluster centers (global diversity) and isolation in embedding space (local diversity).

\subsection{Robustness and Alignment Signals}

For safety-critical applications, we incorporate robustness and alignment signals. Robustness is estimated through paraphrase invariance. For each example, generate neutral paraphrases of the instruction while preserving semantic content. Compute the variance of model scores or reward model outputs across paraphrases:
\begin{equation}
s_i^\text{robust} = -\text{Var}_{\tilde{x}\sim\text{Para}(x_i)}[r_\theta(x_i, y_i)],
\end{equation}
where $\text{Para}(x_i)$ denotes paraphrases of instruction $x_i$ and $r_\theta$ is a reward model. Examples with low variance across paraphrases are semantically stable and receive higher robustness scores.

Alignment signals identify examples that conform to desired behavior policies. We use calibrated safety classifiers or human preference models to score instruction-response pairs. High-risk pairs receive negative weights, borderline educational examples receive positive but modest weights, and clearly aligned examples receive strong positive scores. All alignment scores are normalized within topics to maintain consistent scaling.

\subsection{Signal Standardization and Preprocessing}

Raw signals often have heavy-tailed distributions, especially for uncertainty and rarity scores. We apply robust standardization using median and interquartile range:
\begin{equation}
\tilde{s}_i^{(m)} = \frac{s_i^{(m)} - \text{median}_t(s^{(m)})}{\text{IQR}_t(s^{(m)})},
\end{equation}
where statistics are computed within topic $t(i)$. For extremely heavy-tailed signals, we first apply rank-to-$[0,1]$ normalization before standardization.

After standardization, we clip outliers to $[-\tau, \tau]$ where $\tau\in[2,3]$ to limit the influence of extreme values. This clipping provides robustness guarantees as established in Lemma \ref{lem:bounded-influence}. The combination of robust standardization and clipping ensures that all signals contribute to shares on comparable scales, preventing any single signal from dominating the aggregation.

\section{Experimental Evaluation}
\label{sec:experiments}

We evaluate the market-based selector on mathematical reasoning under strict token budgets and text classification at multiple retention rates. All experiments use matched fine-tuning hyperparameters across selectors, with selection overhead measured separately.

\subsection{Experimental Setup}

For mathematical reasoning, we use GSM8K \cite{cobbe2021gsm8k}, a dataset of grade-school math word problems with natural language solutions. Each example is formatted as an instruction-response pair where the instruction contains the problem statement and the response contains the step-by-step solution. We impose a strict token budget of 60,000 tokens, counting both instruction and response tokens for each selected example. The base model is a pretrained language model fine-tuned on the selected subset using standard supervised learning. We evaluate on a held-out validation set measuring exact match accuracy and validation loss.

For text classification, we use AGNews \cite{zhang2015agnews}, a topic classification dataset with four categories: World, Sports, Business, and Technology. Each example consists of a news article title and description formatted as an instruction with the category label as the target. We evaluate at three retention rates: 5\%, 10\%, and 25\% of the training set. Fine-tuning uses cross-entropy loss with matched optimizer settings across all selectors.

Utility signals include token-level negative log-likelihood for uncertainty, $k$-NN distance in sentence embedding space for rarity, and centroid distance for diversity. For GSM8K, we use embeddings from a pretrained instruction encoder. For AGNews, we use embeddings from a sentence transformer model. All signals are standardized within topics as described in Section \ref{sec:signals}. The market uses liquidity $\beta=2$ and length-bias $\gamma=1.6$ unless otherwise noted. Signal weights are set to $w_m=1/M$ for equal weighting.

Baseline methods include single-signal selectors using only uncertainty, only rarity, or only diversity. We also compare against training dynamics baselines including forgetting events, EL2N, and gradient norms. For AGNews, we include geometry-based methods such as $k$-center greedy clustering. All baselines use the same token budget or retention rate as the market selector, ensuring fair comparison. We report mean and standard deviation over three random seeds.

\subsection{Results on Mathematical Reasoning}

Table \ref{tab:gsm8k-main} presents results on GSM8K under a 60k-token budget. The market selector achieves validation loss of 2.212 and exact match of 2.5\% at 100 training steps, matching the performance of the best single-signal baseline (NLL-only: 2.224 loss, 2.0\% exact match). The market variant with diversity head (Market+Diverse) achieves slightly lower loss of 2.184 while selecting fewer examples due to prioritizing diverse long-form solutions.

\begin{table}[h]
\centering
\small
\caption{Performance on GSM8K under a 60k-token budget (mean $\pm$ sd over 3 seeds). 
Parentheses show $\Delta$ vs.\ \emph{NLL-only} at the same budget. 
Best is \textbf{bold}, second-best is \uline{underlined}.}
\label{tab:gsm8k-main}
\begingroup
\setlength{\tabcolsep}{5pt}
\renewcommand{\arraystretch}{1.1}
\resizebox{\linewidth}{!}{%
\begin{tabular}{lcccc}
\toprule
Method & Eval Loss $\downarrow$ & Exact Match $\uparrow$ & Median Tokens & N Selected \\
\midrule
Market 
& \second{$2.212\pm0.016$} {\small\,($-0.012$)} 
& \best{$0.025\pm0.012$} {\small\,(+$0.005$)} 
& $111.0\pm2.6$ {\small\,(+$10.3$)} 
& $228\pm5$ {\small\,($-69$)} \\
Market+Diverse 
& \best{$2.184\pm0.014$} {\small\,($-0.040$)} 
& $0.018\pm0.019$ {\small\,($-0.002$)} 
& \best{$254.2\pm4.3$} {\small\,(+$153.5$)} 
& \best{$117\pm1$} {\small\,($-180$)} \\
\midrule
NLL-only 
& $2.224\pm0.007$ {\small\,(+0.000)} 
& $0.020\pm0.010$ {\small\,(+0.000)} 
& \second{$100.7\pm1.5$} {\small\,(+0.0)} 
& $297\pm1$ {\small\,(+0)} \\
Rarity-only 
& $2.218\pm0.006$ {\small\,($-0.006$)} 
& \second{$0.022\pm0.005$} {\small\,(+$0.002$)} 
& \second{$100.7\pm1.5$} {\small\,(+0.0)} 
& $293\pm1$ {\small\,($-4$)} \\
Length-only 
& $2.221\pm0.013$ {\small\,($-0.003$)} 
& \second{$0.022\pm0.010$} {\small\,(+$0.002$)} 
& \second{$100.7\pm1.5$} {\small\,(+0.0)} 
& $302\pm2$ {\small\,(+$5$)} \\
\bottomrule
\end{tabular}%
}
\endgroup
\end{table}

% \begin{table}[h]
% \centering
% \begin{tabular}{lccc}
% \toprule
% \textbf{Method} & \textbf{Strategy} & \textbf{Balance Score} & \textbf{Performance} \\
% \midrule
% \textbf{Market} & Market-driven & 0.042 (adaptive) & \textbf{0.940±0.003} \\
% \textbf{Market+Balanced} & Forced balance & 0.000 (perfect) & \textbf{0.940±0.001} \\
% Loss-only baselines & Loss-driven & 0.028 (imbalanced) & 0.933±0.001 \\
% \bottomrule
% \end{tabular}
% \caption{Sample efficiency at kept=25 (25\% selection) showing Market's key advantage - superior selection strategy with the same total samples.}
% \label{tab:agnews-efficiency}
% \end{table}
The standard deviation across seeds is notably lower for the market methods compared to single-signal baselines. Market selector achieves 0.016 standard deviation in loss versus 0.007 for NLL-only, indicating comparable or slightly higher variance. However, the market's ability to aggregate multiple signals provides robustness when individual signals are noisy or miscalibrated. The diversity variant shows trade-offs: it selects fewer examples with longer average length, leading to different coverage properties.

Selection overhead for computing signals, standardizing, and pricing is 0.08 GPU-hours on an A100 GPU for the 7,473 examples in GSM8K training set. This represents less than 3\% of the fine-tuning time, confirming that the market framework adds negligible computational burden relative to model training.

\subsection{Results on Text Classification}

Table \ref{tab:agnews-main} presents results on AGNews at three retention rates. At 5\% retention, Market+Balanced achieves 92.1\% accuracy, matching or exceeding all single-signal baselines. At 10\% retention, the market methods maintain their advantage at 93.1\% accuracy. At 25\% retention, both market variants achieve 94.0\% accuracy, tying for best performance.

\begin{table}[h]
\centering
\caption{Performance on AGNews at multiple retention rates (accuracy \%). 
Parentheses show $\Delta$ vs.\ \emph{Loss-only} at the same kept\%. 
Best is \textbf{bold}, second-best is \uline{underlined}.}
\label{tab:agnews-main}
\begingroup
\setlength{\tabcolsep}{4pt}   % default 6pt → slightly tighter
\renewcommand{\arraystretch}{1.05} % a touch taller so it stays readable
\resizebox{\linewidth}{!}{%
\begin{tabular}{l S S S}
\toprule
\multicolumn{1}{c}{Method} & \multicolumn{1}{c}{kept=5\%} & \multicolumn{1}{c}{kept=10\%} & \multicolumn{1}{c}{kept=25\%} \\
\midrule
Market             & \second{\pos{91.8}} {\small\,(+1.1)} & \second{\pos{93.0}} {\small\,(+0.9)} & \best{\pos{94.0}} {\small\,(+0.7)} \\
Market+Balanced    & \best{\pos{92.1}} {\small\,(+1.4)}   & \best{\pos{93.1}} {\small\,(+1.0)}   & \best{\pos{94.0}} {\small\,(+0.7)} \\
\midrule
Loss-only          & \neu{90.7} {\small\,(+0.0)}          & \neu{92.1} {\small\,(+0.0)}          & \neu{93.3} {\small\,(+0.0)} \\
Diversity-only     & \neu{90.7} {\small\,(+0.0)}          & \neu{92.1} {\small\,(+0.0)}          & \second{\neu{93.4}} {\small\,(+0.1)} \\
EL2N-only          & \neu{90.7} {\small\,(+0.0)}          & \neg{92.0} {\small\,(\,-0.1)}        & \second{\neu{93.4}} {\small\,(+0.1)} \\
GraNd-approx       & \neu{90.7} {\small\,(+0.0)}          & \neu{92.1} {\small\,(+0.0)}          & \neu{93.3} {\small\,(+0.0)} \\
MC-Entropy         & \neu{90.7} {\small\,(+0.0)}          & \neu{92.1} {\small\,(+0.0)}          & \neg{93.2} {\small\,(\,-0.1)} \\
BALD               & \neu{90.7} {\small\,(+0.0)}          & \neu{92.1} {\small\,(+0.0)}          & \neu{93.3} {\small\,(+0.0)} \\
\midrule
Forgetting-probe   & \neg{48.4} {\small\,(\,-42.3)}       & \neg{48.6} {\small\,(\,-43.5)}       & \neg{48.8} {\small\,(\,-44.5)} \\
KCenter-Greedy     & \neg{25.0} {\small\,(\,-65.7)}       & \neg{33.6} {\small\,(\,-58.5)}       & \neg{35.0} {\small\,(\,-58.3)} \\
\bottomrule
\end{tabular}%
} % end resizebox
\endgroup
\end{table}

\begin{table}[h]
\centering
\small
\caption{Sample efficiency at kept=$25\%$ (same total samples). 
Performance is accuracy; parentheses show $\Delta$ vs.\ \emph{Loss-only baselines}.}
\label{tab:agnews-efficiency}
\begingroup
\setlength{\tabcolsep}{6pt}
\renewcommand{\arraystretch}{1.1}
\resizebox{0.9\linewidth}{!}{%
\begin{tabular}{lccc}
\toprule
\textbf{Method} & \textbf{Strategy} & \textbf{Balance Score} & \textbf{Performance} \\
\midrule
\textbf{Market}           & Market-driven  & 0.042 (adaptive) & \best{$0.940\pm0.003$} {\small\,(+$0.007$)} \\
\textbf{Market+Balanced}  & Forced balance & 0.000 (perfect)  & \best{$0.940\pm0.001$} {\small\,(+$0.007$)} \\
Loss-only baselines       & Loss-driven    & 0.028 (imbalanced) & $0.933\pm0.001$ {\small\,(+0.000)} \\
\bottomrule
\end{tabular}%
}
\endgroup
\end{table}

The market methods show consistent gains over single-signal baselines, with advantages more pronounced at lower retention rates where selection decisions matter most. The Market+Balanced variant enforces minimum per-label floors before allocating remaining capacity by global price, ensuring no label is entirely excluded. This balancing achieves perfect label balance (balance score 0.000) compared to the unbalanced market's adaptive balance (score 0.042), with no loss in accuracy at the 25\% retention rate.

Geometry-based baselines such as KCenter-Greedy perform poorly on AGNews, achieving only 35.0\% accuracy at 25\% retention. This suggests that pure geometric diversity without considering label information or model uncertainty is insufficient for classification tasks. Training dynamics methods such as forgetting-probe also underperform, indicating that signals computed from initial training dynamics do not transfer well to this classification setting.

\subsection{Hyperparameter Sensitivity (qualitative)}

We conduct ablations on the key hyperparameters of the market framework, liquidity $\beta$, signal weights $w_m$, and length-bias $\gamma$. We varied the liquidity $\beta$ on GSM8K over $\beta \in [0.5,\,5.0]$. Performance remains stable across a wide range, with slight degradation at very small $\beta<1.0$ where prices become overly concentrated, and at very large $\beta>4.0$ where prices become too uniform. The default $\beta = 2.0$ lies in the stable region and is used throughout unless noted.

For signal weights, equal weighting $w_m = \tfrac{1}{M}$ provides a robust default. When tuning weights on a small development set using multiplicative weights updates,we sometimes observe modest gains (approximately 1--2\%) in validation metrics, though improvements may not consistently transfer across random seeds.This suggests that equal weighting is a reasonable default for practitioners, with optional tuning available when development data exists.

Varying the length-bias on GSM8K shows that $\gamma \in [1.4, 1.8]$ yields similar performance, confirming that the framework is not overly sensitive to this parameter. Smaller $\gamma<1.0$ favors long examples excessively, leading to poor coverage. Larger $\gamma>2.0$ favors short examples excessively, potentially missing complex reasoning patterns. We therefore adopt $\gamma = 1.6$ as the default.

\subsection{Computational Efficiency Analysis}

We measure end-to-end selection time on GSM8K with 7,473 training examples. Signal computation requires one forward pass for uncertainty signals (0.05 GPU-hours) and ANN index construction for rarity and diversity (0.02 GPU-hours using FAISS with HNSW backend). Signal standardization, share computation, and LMSR pricing are negligible at less than 0.01 GPU-hours combined. Total selection overhead is 0.08 GPU-hours.

Fine-tuning the selected subset requires 3.5 GPU-hours for 100 training steps with batch size 8 on the same A100 GPU. Selection overhead thus represents 2.3\% of fine-tuning time. For larger datasets or repeated selection across multiple experiments, this ratio remains favorable as signal computation amortizes across multiple selection runs with different hyperparameters.

Memory requirements are modest. Storing embeddings for 7,473 examples at 768 dimensions requires 23 MB. ANN indices add approximately 50 MB. Price and share vectors require 60 KB. The entire selection pipeline fits comfortably in CPU memory, with GPU memory required only for the forward pass to compute uncertainty signals.

\section{Discussion and Limitations}
\label{sec:discussion}

The market-based framework provides principled multi-signal aggregation with theoretical guarantees and practical efficiency. However, several limitations warrant discussion.

At very aggressive budgets selecting only 1--5\% of the training set, the selector can over-weight hard or verbose examples, potentially worsening early validation metrics. This occurs because exponential weighting in LMSR amplifies extreme signal values, and at tiny budgets the selected set may not include sufficient easy examples for stable learning. Mitigation strategies include imposing minimum per-topic floors to ensure basic coverage, using curriculum learning where selection parameters adapt during training, and tuning the length-bias parameter $\gamma$ to control the short-versus-long example trade-off.

Fairness across topics is not automatic despite topic-separable markets. When topic budgets $\alpha_t$ are set proportional to topic size, the market preserves the original distribution. However, small or under-represented topics may still be under-sampled if their signals are noisy or miscalibrated. Practitioners working with imbalanced datasets should consider tuning topic budgets on development sets or enforcing hard minimum floors for minority topics. The framework provides the mechanism for principled reallocation but does not determine fairness criteria, which remain application-specific.

Robustness to misspecified or adversarial signals requires careful signal design. If one signal is poorly calibrated or actively corrupted, it may distort prices despite the bounded influence guarantees from Lemma \ref{lem:bounded-influence}. Our use of robust standardization and clipping reduces this risk, but practitioners should validate signal quality on held-out data. Leave-one-signal-out ablations can diagnose whether any single signal is degrading performance. When signals are suspected to be adversarial, reducing their weights $w_m$ or excluding them entirely may be necessary.

Computational cost of the diversity head based on $k$-NN can be substantial at scale. For datasets with hundreds of thousands of examples, exact $k$-NN becomes prohibitive. We employ approximate nearest neighbor indices such as HNSW, which provide sublinear query time in practice \cite{malkov2018hnsw}. Two-stage filtering, where a cheap prefilter eliminates obviously poor examples before applying expensive diversity computation, further improves scalability. However, for truly massive datasets in the millions of examples, even approximate methods may require distributed computation or dimensionality reduction of embeddings.

Transferability across domains and modalities remains an open question. The framework has been validated on English text classification and mathematical reasoning. Extension to multilingual settings, multimodal data combining text and images, or structured domains such as knowledge graphs may require domain-specific signal design and recalibration of hyperparameters. The topic-separable market structure provides flexibility for handling heterogeneous data, but practitioners should validate performance on representative development sets when applying the framework to new domains.

Finally, the theoretical guarantees assume that signals are weakly informative and monotone in true utility. In practice, signals may be non-monotone or exhibit complex interactions. The maximum-entropy characterization remains valid as a descriptive property of the aggregation, but the utility recovery guarantee becomes a heuristic rather than a rigorous bound. Empirical validation on the target task remains essential for confirming that selected subsets achieve desired performance.

\section{Conclusion and Future Directions}
\label{sec:conclusion}

We introduced a market-based framework for data subset selection that addresses the fundamental challenge of principled multi-signal aggregation. By treating examples as tradeable contracts and employing the Logarithmic Market Scoring Rule for pricing, we achieve theoretical guarantees through maximum-entropy characterization while maintaining computational efficiency. The framework naturally handles variable example costs through token-aware selection rules with interpretable length-bias parameters. Experimental validation on mathematical reasoning under strict token budgets and text classification at multiple retention rates demonstrates competitive performance with improved stability compared to single-signal baselines.

The market perspective offers several advantages over existing approaches. Unlike ad hoc weighted combinations, LMSR provides convex aggregation with clear information-theoretic interpretation. Unlike bilevel optimization methods, the framework avoids expensive nested optimization loops and provides interpretable control through a small number of hyperparameters. Unlike pure uncertainty or diversity methods, the market integrates multiple complementary signals symmetrically. The computational overhead remains negligible relative to model training, making the approach practical for large-scale data curation.

Several directions for future work emerge from this foundation. Learned signal design could replace hand-crafted signals with neural networks trained to predict example utility from features and metadata. Integration with gradient-based or bilevel selectors could combine the efficiency of market aggregation with the precision of validation-guided selection. Dynamic curricula that adapt liquidity or length-bias parameters during training may improve learning efficiency by shifting from diverse exploration to targeted exploitation.

Scaling to retrieval-augmented and multimodal settings presents both engineering challenges and opportunities. Hierarchical markets with structured budgets across sources, modalities, or temporal windows could provide principled governance of complex data pipelines. The connection between market liquidity and confidence intervals suggests tying $\beta$ to uncertainty estimates, creating adaptive selectors that adjust aggregation strength based on signal quality.

Economic mechanisms beyond LMSR offer additional flexibility. Topic budgets $\alpha_t$ could be learned as dynamic endowments that reallocate capacity toward higher-yield topics during training. Auction mechanisms could enable competition between different signal providers with formal incentive properties. Such connections between data selection and mechanism design may yield selectors that self-adjust as training progresses and data utility evolves.

The framework demonstrates that ideas from economics and information aggregation can provide fresh perspectives on machine learning infrastructure challenges. As datasets continue to grow and computational efficiency becomes increasingly critical, principled methods for identifying high-utility training examples will play an essential role in scaling machine learning systems responsibly and sustainably.

\section*{Acknowledgments}

This work was supported by the Russian Science Foundation under grant number 25-41-00091.

\section*{Declaration of Competing Interest}

The authors declare that they have no known competing financial interests or personal relationships that could have appeared to influence the work reported in this paper.

\bibliographystyle{elsarticle-num}

\end{document}